# Real Time Object Detection System with YOLO and CNN Models: A Review

Viswanatha V[1*]
*Asst.Professor, Electronics and Communication Engineering Department,
Nitte Meenakshi Institute of Technology, Bangalore.
Email – viswas779@gmail.com*

Chandana R K[2]
*MTech student, Electronics and Communication Engineering Department
Nitte Meenakshi Institute of Technology, Bangalore. Email- chandanark99@gmail.com*

Ramachandra A.C[3]
*Professor and Head, Electronics and Communication Engineering Department
Nitte Meenakshi Institute of Technology, Bangalore. Email- ramachandra.ac@nmit.ac.in*

**Abstract-** The field of artificial intelligence is built on object detection techniques. YOU ONLY LOOK ONCE (YOLO) algorithm and it's more evolved versions are briefly described in this research survey. This survey is all about YOLO and convolution neural networks (CNN) in the direction of real time object detection. YOLO does generalized object representation more effectively without precision losses than other object detection models. CNN architecture models have the ability to eliminate highlights and identify objects in any given image. When implemented appropriately, CNN models can address issues like deformity diagnosis, creating educational or instructive application, etc. This article reached at number of observations and perspective findings through the analysis. Also it provides support for the focused visual information and feature extraction in the financial and other industries, highlights the method of target detection and feature selection, and briefly describe the development process of yolo algorithm.

## I. INTRODUCTION

Object detection strategies are the establishment for the engineered insights subject. This paper offers a brief evaluation of the you only look once (YOLO) algorithm of rules and another prevalent variation. Through the assessment, it is understood that numerous comments and proper results show the variations and likenesses between Yolo version and between Yolo and convolutional neural networks (CNNS). The relevant perception is the Yolo algorithm development continues to be Ongoing. This article in brief depicts the improvement procedure of the Yolo set of rules, summarizes the methods of goal Recognition and characteristic choice, and provides literature assistance for the centered image news and characteristic extraction as shown in fig.1 [1].

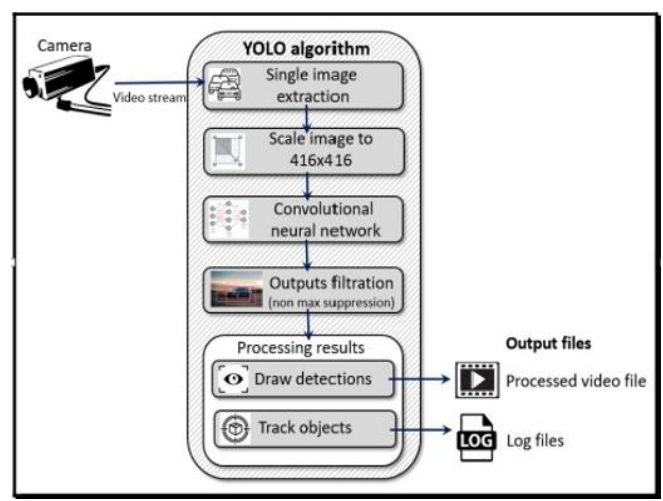

Figure.1

YOLO is a new approach of object detection. The classifiers are used to detection in earlier works in an object detection. It is consider here that frame object detection as a regression issue to spatially separated bounding package container and associated class probabilities as an alternative object detection as shown in fig.2.

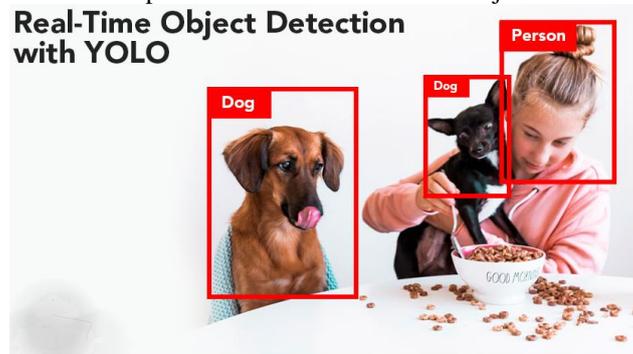

Figure.2

In a single evaluation, a single neural network predicts a boundary bins and class possibilities directly from complete images. The entire direction pipeline may be optimized start to finish because it is a single network. On the effectiveness of detection without delay a unified architecture moves incredibly quickly and as well 45 frames per 2D frame yolo model process image in real time. Yolo, a scaled down version of the network, achieves double the coverage of the other real time detectors while operating at an incredible 125 frames per second. Yolo makes more localization errors than modern detection algorithm but is less likely to anticipate false positives based on past data. Yolo eventually picks up and highly preferred representation of items while generalizing from natural photographs to other domain names like painting, its outperform competing detection techniques like DPM and R-CNN [2].

II. Approaches of implementation of YOLO and CNN

Object detection is a technology that detects the semantic Objects of a category in virtual snap shots and films. Certainly, one of its Actual-time packages is self-riding automobiles. In this, our challenge is to locate a couple of items from a photo. The maximum common Item to come across on this utility is the car, motorcycle, and Pedestrian. For locating the gadgets within the photograph, Object localization and should find multiple item in real-time structures. There are various techniques for item Detection, they can be break up into classes, first is the Algorithms primarily based on classifications. CNN and RNN come below this category. On this, pick out the involved Regions from the picture and ought to classify them the use of Convolutional neural community. This technique may be very sluggish Due to the fact it should run a prediction for every decided on Vicinity. The second one class is the algorithms primarily based on Regressions. Yolo approach comes below this category. In This, it might not choose the fascinated regions from the photograph. Instead, it expect the training and bounding containers of the complete picture at a single run of the algorithm and detect a couple of gadgets using an unmarried neural community. Yolo Set of rules is rapid as compared to other classification Algorithms. In actual time our algorithm technique 45 frames consistent with 2d. Yolo algorithm makes localization mistakes however Predicts less fake positives in the background [3].

Humans can without problems come across and pick out gadgets of their environment without attention in their instances, irrespective of what position they're in and whether or not they are the wrong way up, one-of-a-kind in shade or texture, in part occluded, and so forth. To extract information about the objects and shapes in a picture, the same item detection and recognition on a computer requires a lot of processing. Identifying an object in an image or video is referred to end as an object detection in computer vision. The main steps in object detection or future extraction, which is imported for surveillance, cancer reduction, car identification, and underwater object detection, among other applications. Different approaches had been used to accurately and correctly identify the object for specialized packages. These suggested solutions still have an issue with inaccuracy and inefficiency, though. Device learning and deep neural network approaches are more successful in addressing these issues of object detection [4].

Convolution neural networks CNN for widely used in image processing and other areas with the development of artificial intelligence because of their fantastic performance. However, because it's a set of rules is so complex computationally, CNN faces enormous difficulties capturing the interest of mobile devices. FPGA becoming ideal choices for CNN deployment due to their advantages of high performance, reprogramming, and seldom energy usage. The yolo algorithm views target detection as a regression problem in comparison to other CNN algorithms.

It is a one-step algorithm that runs quickly and requires little calculation. It is much more appropriate for FPGA hardware structure. The issue of massive computational of CNN and constrained assets on FPGA chip is resolved, and the parallelism of FPGA is used to boost the CNN. This is accomplished by improving the Yolo network version and fixed point, among other things. Experimental finding demonstrate that the strategy suggested in this paper as significantly reduced the operating cost while maintaining accuracy and as an essential reasonable cost within the creation of mobile terminals and real time computing. In order to speed up the deep learning yolo community, this paper offers an advanced set of guidelines for deep learning Yolo network that are entirely based on Xilinx FPGA. By using the parallelism function of the FPGA to speed up the CNN, the acceleration set of rules specifically addresses the issue that the CNN requires a significant amount of calculation while the FPGA has limited resources on the chip. The implementation of this set of rules specifically includes all three parts: the first one is the FPGA Yolo community design the second one is the selection of the activation characteristics and the last one is 16 bit constant point optimization of the weight parameter [5].

Development plan is presented as a result of the lack of common accuracy and miss detection in the way of actual roots in gold detection through the Yolo V3 community in order to investigate the anchor range and component ratio, the original network clustering set of rules is updated using the ok-means clustering algorithm similarly, so that you can enhance the performance of the road target detection set of rules, the present network output is upgraded, and a 104*104 feature detection layer is brought, and the feature map output through 8 instances sampling can be output through 2 instances up sampling, and 4 the characteristic maps of down-sampling are stitched together, and the 104*104 feature maps received can efficiently lessen the disappearance of features. Via the experimental effects, it is able to see that compared with the stepped forward yolov3 set of rules, the common detection accuracy of the stepped forward set of rules for road goal detection is quite excessive to a few.17%, and the missing detection rate is reduced by 5.62% [6].

In this paper author proposed laptop vision functions of onboard cameras and embedded devices are widely employed with general unmanned aerial vehicles (UAV). However ,it is very difficult program to study the real time seen through the object identification approach go to the limited memory and computing capability of embedded devices at the UAV platform. This study evaluates the performance of different you yolo collection methods on the Pascal voc dataset, employees map and frame per second at assessment measures, and applies the learning outcomes to the xtdrone UAV simulation platform for testing in order to overcome those difficulties. The map of Yolo V4 is 87.48%, which is 14.2% better than that of yolo V3 according to our evaluation of Yolo V3, Yolo V3 tiny, and yoov3-spp3 and yolov4-tiny on the pascal voc benchmark dataset. Frame per second reaches 72, and the test sets check time is one or 103.8 seconds, the validation sets shortest test time is Yolo V4, but the map only achieves 50.06% accuracy which is less precise than Yolo V3 tiny. Using simulation on the xtdrone platform to analyze the performance of five modules on the Pascal voc data set come on this paper ultimately comes to the conclusion that Yolo V3 can match the demands of real time, lightweight, and excessive precision [7].

The main objective of real time object detection is to locate the location of an object in a supply picture accurately and mark the item with the appropriate category. In this paper it used actual time object detection you appearance best as soon as (yolo) set of rules to train our device studying version. Yolo is a clever neural network for doing object detection in actual time and with the assist of coco dataset the set of rules is trained to perceive one of a kind items in a specific picture. After training this method locate the object in actual time with ninety% accuracy.

The aim of item detection is to come across all times of gadgets from an acknowledged magnificence, consisting of humans, motors or faces in an image. Commonly, most effective a small quantity of instances of the object are gift inside the picture, however there are a very huge range of possible places and scales at which they could occur and that want to by some means be explored. Each of the detection is mentioned with some form of pose data. This could be as simple because the region of the object, an area and scale, or the volume of the object described in terms of a bounding field. In different conditions, the pose facts is extra specific and includes the parameters of a linear or nonlinear transformation. As an example, a face detector may additionally compute the places of the eyes, nose and mouth, further to the bounding box of the face [8].

Synthetic intelligence is being tailored by the sector on the grounds that past few years and deep gaining knowledge of performed a crucial position in it. This paper focuses on in depth learning and how it is used to find and tune the devices. Deep learning uses algorithms that are impacted by the structure and functions of the brain. Working with such algorithms as the advantage that performance generally improves with growth in data, as opposed to conventional learning algorithms whose performance stabilizes regardless of boom in the amount of data a number of most popular research topics in computer vision application have put real time item monitoring at the forefront. Despite the impressive progress made at this paper, it is still difficult to effectively and accurately track the objects in real time at a significant level. The capabilities of images and movies for security and oversight packages are extracted while defining detection and monitoring algorithm. You only look once (Yolo), location based fully convolutional neural network (RCNN), and quicker samples of popular object detection algorithm (f-RCNN). RCNN is more accurate than other algorithms, but yolo outperforms them when speed is prioritized over accuracy. Yolo treats object detection as a regression issue and provides elegance chances for found images.

Real world circumstances drive the need to use the Yolo algorithm to find object entities base. When pace is taken into account as if performance indicator, Yolo outperforms other object detection algorithms. Regression is used by the community to predict the bounding containers and classes of entities in a single pass of algorithm, as opposed traversing the network more than once to identify and output the opportunity of the elegance. This gives it a real and noticeable advantages over earlier technology in terms of increased frame per second. When compared to other performances like CNN and its various version this result in an object detector using Yolo V3 with impressive overall performance metrics, including accuracy, precision, and everything it is different from other. This enables the builders to use the Yolo V3 for real time monitoring packages in order to benefit from its performance improvements over its antiquated counterparts [9].

By using photographic records enhancement, it is possible to expand the pool of records without actually adding any new information. In this study, researchers assessed the effectiveness of image manipulation techniques like the double phantasm and inversion on the overall performance of the face detection for information enhancement needs. The study found that the limited amount of data available to create a teaching version of the facts was a weakness in the facts that were acquired. The goal of this people like research paper is to increase the variety of the facts such that come up and gives various comparable datasets, it can make prediction with accuracy. Yolo V3 is used to perform face detection on photos, and accuracy results from the dataset and before including records are compared.

Object detection as received a lot of attention in the field of computer vision carried out in various areas, including the clinical field, pastime reputation, security tracking in agencies, and automated manage robots. Famous item detection techniques, before everything, had been recognized by way of the use of feature extraction techniques together with histogram of orientated gradients (hog), speeded-up strong features (surf), nearby binary styles (lbp), and additionally coloration histograms. The system of taking pictures the characteristics of gadgets that could describe the traits of items is the number one approach in the function extraction technique [10].

A straightforward, ready-to-use, unified object detection model that works with complete images. YOLO also generalized well to unused spaces utilized in applications that depend on quick, strong object detection. A degenerative show built for recognizing degraded images like obscured and loud pictures has the model being prepared with these debased pictures. This model performed superior in terms of detecting degraded pictures and adapted way better with complex scenes. For discovery shallow person on foot highlights, a YOLO v2 organize was adjusted by including three Pass through layer to them. The number of detection frames can reach 25 frames per second, which meets the demands of real-time performance. To recognize indoor impediments an unused strategy of using profound learning together with a light field camera was utilized. The strategy recognizes the deterrents and perceives its information. YOLO connected to vehicle entryway board welding panel lines can distinguish and distinguish patch joint accurately the calculation can too identify the position of the solder joints and more [11].

Object Detection has received a lot of research attention in recent years because of its tight association with video analysis and picture interpretation. The foundation of conventional object detection technique is shallow trainable structures and handmade features. Building intricate ensembles that incorporate several low-level picture features with high-level context from object detectors and scene classifier can readily stabilize their performance. In order to solve the issues with traditional architectures, more potent tools that can learn semantic, high-level and deeper features are being offered as a result of deep learning's quick development. In terms of network architecture, training methodology, optimization function, etc., these models behave differently. In this paper, it explore object detection frameworks based on deep learning. A brief background of deep learning is provided before our review. It brief history of deep learning and its illustrative tool, the convolutional neural network (CNN), is given before our review. Then it concentrate on common generic object detection architectures with a few changes and helpful tips to further enhance detection performance.

This paper also provides overview of a number of specific task, such as salient item detection, face detection and pedestal detection, as different specific detection task exhibit different characteristics. To evaluate different approaches and get some insightful results experimental analysis are also offered. In order to provide direction for future work object identification and pertinent neural network based learning systems, a number of promising directions and tasks are provided [12].

An aspect of your computer vision that is constantly in use is object detection. Object identification has made remarkable progress, from the first manual future extractor to the future extractor dominated by deep convolution network. Researchers started paying more attention to CNN in 2012 when Alexnet won the title image net with significant benefits (convolutional neural report). CNN has made remarkable success since 2012 in many facets of computer science, and object vision detection is no exception. As of now, two step end single step CNN-based object detection network can be loosely split into two categories. The two stage object detection network typically consists of three stages: The first stage involves creating multiple delimitations for the proposal cans; the second stage involves performing the object classification. R-CNN, Fast R-CNN, SPPNet,R-CNN faster, FPN, and Mask-RCNN are some of the representative network for the prediction of position information on the bounding boxes of the proposition. The single phase object detection network is an end to end model, meaning that only one network inference is required from the input image to the network output. Its representative objectives include the YOLO series, SSD and RetinaNet. The benefits and drawbacks of the two strategies mentioned above very. Although the single stage detector is faster than the two stage one, the location accuracy of the two stage detector is marginally higher. However, there is a significant issue for practical application computationally expensive and memory consuming. This problem exists independent of the type of detecting network [13].

In the field of goal recognition, YOLO algorithm has carried out well. In this paper, it enhance the brand new YOLO community version via way of means of converting the residual devices to dense connection in the CSP module and including channel interest mechanism. The advanced community version alleviates the vanishing-gradient hassle, complements function propagation, encourages function reuse, and decreases the quantity of parameters. What's more, it can adaptively recalibrate the channel records of the function maps and enhance the overall performance of goal detection. Experimental consequences display that the advanced YOLO community version greatly improves the detection accuracy. In addition, it optimizes the hassle of lacking and miss-detecting targets convolutional neural network (CNN) has end up the famous deep gaining knowledge of set of rules for goal recognition. However, while the function maps attain the stop of the community structure after a couple of convolutions, the function facts in it may disappear. To clear up this type of problem, pupils placed forward a few answers and fashions, consisting of ResNet and DenseNet . Each of those models has a key function that they devise a short route among the front community layer and the again community layer. In the ResNet community model, channels are brought up, but channels are linked within the DenseNet community model.

    It guarantees that the facts go with the drift among layers within the community is maximized. And as it does now no longer want to analyze redundant function maps, every layer community can attain extra function facts with fewer parameters. In addition, DenseNet's facts go with the drift and gradients throughout the community are improved, in order that it turns into less difficult to be trained [14].

Convolution neural networks [CNN] were widely used in photographs and other parts as artificial intelligence (AI) advanced due to their superior performance. Although CNN faces significant challenges in the development of mobile devices due to its computationally complex set of algorithms. FPGA are good candidates for CNN deployment because of their high performance, re-programmability, and sporadic energy usage. The YOLO set of rules views goal detection as a regression issue in comparison to various CNN algorithms. It is a one-step set of rules that execute quickly and require little calculation it is suitable for hardware platform to implement. This paper suggests a sophisticated set of guidelines for a deep learning YOLO community that are entirely based on Xilinx ZYNQ FPGA. The issue of large computational of CNN and confined assets on FPGA chips as resolved by improving the YOLO community version and fixed point, etc. The parallelism FPGA is then exploited to increase the seat and length. Results from experiments show that the method suggested in this study as greatly improved operational cost while maintaining accuracy and as a very reasonable cost inside the consideration of CNN real time computing [15].

In this paper, the author proposed a method to classify human and automotive objects by integrating a support vector machine (SVM) with the deep learning version, you only look once (Yolo), in a high resolution automotive radar device. The constraints of gold envision from the yellow version or incorporated to the SVM in order to improve the overall performance of the class. The result from the yolo and SVM may then be combined with the predetermined target bonds to advance the overall class accuracy. The results showed that they suggested method outperforms those obtained using yolo or in terms of overall performance.

    A new device version that combines the conventional method, an SVM, and the deep learning version, you only look once (yolo), to categorize people and motors. Based on real measurements, the proposed technique can classify people and motors with over 90% accuracy in high decision automotive radar will stop using the suggested version, the SVM and yolo version are training using range angle (RA) domain information from preprocessed radar signal to classify the same scene. In particular, the enhanced class performance by utilizing boundary containers from the yolo version. Finally, by integrating every result, the overall class accuracy is increased [16].

In this paper the author describes the CPU based yolo, a real time object detection version that may be used with non GPU computers to benefit users of low configuration computers. Many well developed algorithms are available for object detection including yolo, Faster R-CNN, Fast R-CNN R, R-CNN, mask R-CNN, R-FCN, SSD, Retinanet, and others. YOLO is a set of deep neural network rules for object detection that is more accurate and quick than most other algorithms. YOLO is made of four computers that are entirely GPU based and media graphics card with at least 12 GB of memory. In this paper they are implemented yolo with open CV in order to make real time object detection possible on CPU based computers. On several non GPU computers, our version accurately and reliably detects an object from video at a frame rate between 10.12 and 6.29 frame per second. The mAP for CPU based Yolo is 31.5% the evaluation of computer vision using deep learning as being built up and completed throughout time mostly using one specific, well known set of rules called convolution neural networks. It has a convolution layer where the output is produced by convolving a clear output with a variety of input components. The environment of a convolution layer allows for the extraction of related styles from companion input as shown in fig.3. Furthermore, compared to a tightly linked layer, a convolution layer tends to require a considerably less weight to be determined because filter are not used in convolution layer [17].

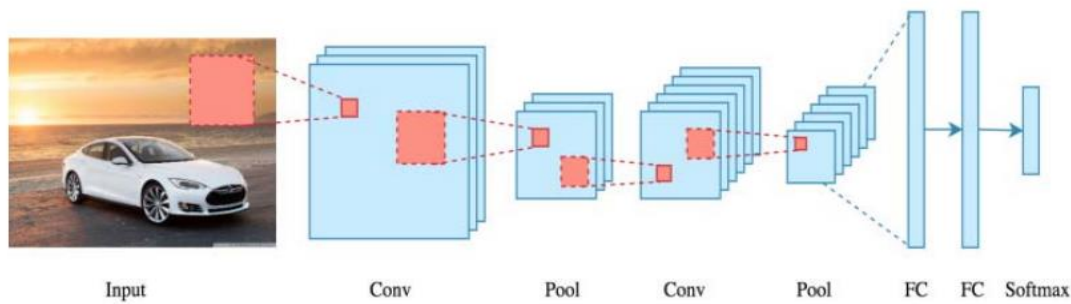

Figure.3 Convolutional Neural Networks design

This study builds a smart monitoring device in response to the current status of the monitoring device, which requires human management and isn't intelligent enough. The device completes the evaluation of the pictures content, implement smart monitoring, and uses deep learning to distinguish things. The program of the smart monitoring gadget is realized using the twisting moment as an example. The yolo V2 version curling automated snapping images device detection record set throughout the test had a precision rate of 94.3%, and the detection speed was 13 frame per second. The yolo version can accommodate the curling the photo taking devices real time requirements. That equipment completed its role of monitoring during the actual photographic process. For the first time, it is applied in this paper to curling gold detection underneath the curling smart monitoring equipment the digital camera platform is operated at the same time based on Yolo V2 curling detection result. The smart monitoring of the curling is completed and combined with this straightforward recording of the curling track. Study shows that yolo V2 fully satisfies the real time requirements with detection accuracy for curling goals of 94.3% and detection rate of 30 frames per second. The yolo based fully curling smart monitoring equipment can guarantee accuracy and real time performance at the same time as well as complete the actual photographic task. The content smart assessment model of the digital camera previews is introduced by a device architecture of the curling smart monitoring device. The teaching of Yolo apply to the smart evaluation version is introduced in the third element along with the evaluation of outcomes [18].

Real-time detection of the pavement surroundings is an essential if part of independent riding technology. This paper gives a real time automobile detection device primarily based totally on embedded devices. Based on the prevailing YOLOv3-tiny neural community shape, this paper proposes a brand new neural community shape the YOLOv3- tiny, and quantify the community parameter in the community. Trading complexity of competing in embedded devices, making the proposal neural community shape more appropriate for embedded devices. The new shape is tested, earlier than quantization the YOLO V3 live's detection precision way of means of the convolution operation and dispatched to the Yolo layer for goal detection. At the same time, the 13x13 function map is merged with the 26x26 function map is dispatched second one yolo layer for goal detection. The entire network shape makes use of yolo layers to hit upon exceptional size of function graphics, in order that the community version can hit upon last targets in addition to small targets. The community shape of YOLO V3 tiny can gain 87.79% mAp at the take a look at set, after quantization of community parameters can gain 69.79% mAP and detection pays can gain 28 FPS. The layer shape of Yolo V3 that is specifically composed of convolution layer and pool layer as shown in fig.2. A small function map of 13x13 is received via way of means of the usage of pooling layer to carry out 5 down sampling operation at the input picture, after which the function is extracted [19].

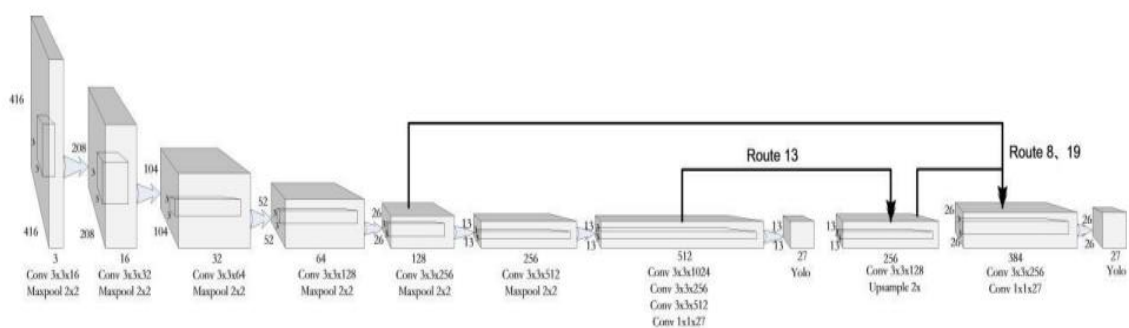

Figure.4 The structure of YOLOv3-tiny network

Breed type is type of detected animal with the aid of using the usage of the pc imaginative and prescient method. This observe is accomplished to assess the overall performance of livestock detection and breed type consistent with particular areas at the frame of livestock that differ consistent with their breed. You Only Look Once (YOLO) the set of rules is used for livestock detection and breed type on the information set. A unique information set is generated using images gathered from Google images in order to demonstrate the usefulness of the suggested strategy in the detection and type of livestock. Experimental results show that the YOLO set of rules can accurately identify the breed of animals on autograph with 92.85% accuracy. With the use of item detection set of rules YOLOv4, this observation seeks to livestock dictation and type. The item detection set of criteria, however, provides the detection and type of the cattle in the photograph. YOLO are used to identify livestock and categories breed for this purpose. The data set was manually constructed from images collected from various structure because there may not be a formally educated version and information sector version education, the title said was manually generated using photographs collected from various structures. As it expand this data set, it pay particular attention to the gender of animals like calf, cow, and steer. The data set was manually classified after that. The positioning and sophistication of the creatures inside the image are determined as a result of being this tagging. Later, version education is accomplished with the aid of using feeding the version with the pc imaginative and prescient method used in this observe. The experimental outcomes display that the educated version finished 92.85 accuracy at the given take a look at set [20]-[23].

### III. CONCLUSION

In addition to providing an overview of the real time object detection technique used by YOLO, this paper discusses the core CNN algorithm structure. CNN architecture models have the ability to eliminate highlights and identify objects any given image. When implemented appropriately, CNN models can address issues like deformity diagnosis, creating educational or instructive application, etc. In practice, yolo has a lot of advantages over other CNN algorithm. Yolo can train the complete model in parallel since it is a unified object detection model that is easy to build and train in accordance with its simple loss function. The best speed and accuracy tradeoff for object detection is offered by second major version of YOLO known as YOLOv4. In addition, Yolo is the most advanced object identification technique and is advised for real time object detection since it generalized object representation more effectively than other object detection models. With these accomplishments, it is clear that field of object detection as a bright future.

REFERENCES

1. Peiyuan Jiang, Daji Ergu*, Fangyao Liu, Ying Cai, Bo Ma ; A Review of Yolo Algorithm Developments ;The 8th International Conference on Information Technology and Quantitative Management ( ITQM 2020 & 2021)
2. Joseph Redmon∗ , Santosh Divvala∗, Ross Girshick , Ali Farhadi; "You Only Look Once: Unified, Real-Time Object Detection" ; University of Washington∗ , Allen Institute for AI† , Facebook AI Research,9 may 2016
3. Geethapriya. S, N. Duraimurugan, S.P. Chokkalingam; "Real-Time Object Detection with Yolo" International Journal of Engineering and Advanced Technology (IJEAT) ISSN: 2249 – 8958, Volume-8, Issue-3S, February 2019.
4. Tanvir Ahmad , 1 Yinglong Ma , 1 Muhammad Yahya,2 Belal Ahmad,3 Shah Nazir , 4 and Amin ul Haq; "Object Detection through Modified YOLO Neural Network" Research Article, 6 June 2020.
5. Zhenguang Li; "An improved algorithm for deep learning YOLO network based on Xilinx ZYNQ FPGA"; 2020 International Conference on Culture-oriented Science & Technology (ICCST).
6. JIN Zhao-zhao1, ZHENG Yu-fu1; "Research on Application of Improved YOLO V3 Algorithm in Road Target Detection"; ICMTIM 2020 Journal of Physics: Conference Series.
7. Shuo Wang; "Research Towards Yolo-Series Algorithms: Comparison and Analysis of Object Detection Models for Real-Time UAV Applications"; IoTAIMA 2021 Journal of Physics: Conference Series.
8. Priya Kumari , Sonali Mitra , Suparna Biswas, Sunipa Roy, Sayan Roy Chaudhuri, Antara Ghosal, Palasri Dhar, Anurima Majumder; "YOLO Algorithm Based Real-Time Object Detection"; June 2021| IJIRT | Volume 8 Issue 1 | ISSN: 2349-6002
9. Dr. N. Murali Krishna1 , Ramidi Yashwanth Reddy2 , Mallu Sai Chandra Reddy3 , Kasibhatla Phani Madhav4 , Gaikwad Sudham; "OBJECT DETECTION AND TRACKING USING YOLO "; 2021 Third International Conference on Inventive Research in Computing Applications (ICIRCA) .
10. Rakha Asyrofi, Yoni Azhar Winata ; "The Improvement Impact Performance of Face Detection Using YOLO Algorithm"; Proc. EECSI 2020 - 1-2 October 2020
11. Rekha B. S.1, Athiya Marium2, Dr. G. N. Srinivasan3, Supreetha A. Shetty; "Literature Survey on Object Detection using YOLO"; International Research Journal of Engineering and Technology (IRJET) , Volume: 07 Issue: 06 | June 2020.
12. Zhong-Qiu Zhao, Member, IEEE, Peng Zheng, Shou-tao Xu, and Xindong Wu, Fellow, IEEE; "YOLOv3: An Incremental Improv Object Detection with Deep Learning: A Review"; 16 Apr 2019
13. Yonghui Lu1, Langwen Zhang2, Wei Xie; "YOLO-compact: An Efficient YOLO Network for Single Category Real-time Object Detection"; 2020 IEEE

14. Fengxi Yan, Yinxia Xu; "Improved Target Detection Algorithm Based on YOLO"; 2021 4th International Conference on Robotics, Control and Automation Engineering
15. Zhenguang Li; "An improved algorithm for deep learning YOLO network based on Xilinx ZYNQ FPGA"; 2020 International Conference on Culture-oriented Science & Technology (ICCST).
16. Woosuk Kim, Hyunwoong Cho, Jongseok Kim, Byungkwan Kim, and Seongwook Lee; "Target Classification Using Combined YOLO-SVM in High-Resolution Automotive FMCW Radar"; 2020 IEEE Radar Conference.
17. Md. Bahar Ullah; "CPU Based YOLO: A Real Time Object Detection Algorithm"; 2020 IEEE Region 10 Symposium (TENSYMP), 5-7 June 2020, Dhaka, Bangladesh.
18. Shujun Zhang, Shanzhen Lan, Qi Bu, Shaobin Li; "YOLO based Intelligent Tracking System for Curling Sport"; ICIS 2019, June 17-19, 2019
19. Shaobin Chen1, Wei Lin11; "Embedded System Real-Time Vehicle Detectionbased on Improved YOLO Network";2019 IEEE 3rd Advanced Information Management,Communicates,Electronic and Automation Control Conference (IMCEC 2019)
20. Gizem Nur Uzun, Alperen Yılmaz, Mustafa Zahid Gürbüz, Oğuzhan Kıvrak; "Detection and Breed Classification of Cattle Using YOLO v4 Algorithm"; 2021 International Conference on INnovations in Intelligent SysTems and Applications (INISTA)
21. Viswanatha, V., et al. "Intelligent line follower robot using MSP430G2ET for industrial applications." *Helix-The Scientific Explorer| Peer Reviewed Bimonthly International Journal* 10.02 (2020): 232-237.
22. Viswanatha, V., Ashwini Kumari, and Pradeep Kumar. "Internet of things (IoT) based multilevel drunken driving detection and prevention system using Raspberry Pi 3." *International Journal of Internet of Things and Web Services* 6 (2021).
23. Viswanatha, V., A. C. Ramachandra, and R. Venkata Siva Reddy. "Bidirectional DC-DC converter circuits and smart control algorithms: a review." *Journal of Electrical Systems and Information Technology* 9.1 (2022): 1-29.